# "ONCE UPON A TIME…" LITERARY NARRATIVE CONNECTEDNESS PROGRESSES WITH GRADE LEVEL: POTENTIAL IMPACT ON READING FLUENCY AND LITERACY SKILLS

Marina Ribeiro[1,2], Bárbara Malcorra[1], Diego Pintor[3], Natália Bezerra Mota*[1,4]

[1] Research department at Mobile Brain, Mobile Brain, Rio de Janeiro, Brazil

[2] Bioinformatics Multidisciplinary Environment (BioME), Federal University of Rio Grande do Norte (UFRN), Natal, Brazil

[3] Technology department at Árvore, Árvore, Rio de Janeiro, Brazil

[4] Institute of Psychiatry (IPUB), Federal University of Rio de Janeiro (UFRJ), Rio de Janeiro, Brazil

* Corresponding author

**Abstract**

Selecting an appropriate book is crucial for fostering reading habits in children. While children exhibit varying levels of complexity when generating oral narratives, the question arises: do children's books also differ in narrative complexity? This study explores the narrative dynamics of literary texts used in schools, focusing on how their complexity evolves across different grade levels. Using Word-Recurrence Graph Analysis, we examined a dataset of 1,627 literary texts spanning 13 years of education. The findings reveal significant exponential growth in connectedness, particularly during the first three years of schooling, mirroring patterns observed in children's oral narratives. These results highlight the potential of literary texts as a tool to support the development of literacy skills.

**Main text**

Reading is not an innate human ability; instead, it requires the brain to repurpose neural circuits to decode letters, phonemes, and words (Dehaene et al., 2015). Consequently, acquiring reading skills represents a pivotal milestone in a child's cognitive development (Dehaene et al., 2015), profoundly influencing their broader mental organization and cognitive architecture (Mota et al., 2023).

Vygotsky's Zone of Proximal Development (ZPD) concept defines the gap between what a learner can accomplish independently and what they can achieve with appropriate guidance or collaboration. Gradual exposure to increasingly complex narratives offers a promising approach to enhancing and solidifying reading skills (Amendum et al., 2018). Selecting an appropriate book can create enjoyable reading experiences by aligning with the language patterns familiar to children, making a text of comparable complexity to their abilities an optimal choice. However, applying the ZPD requires a reliable and scalable method for assessing text complexity, ensuring reading materials are well-matched to the learner's developmental stage.

Computational methods based on Natural Language Processing (NLP) are innovative for measuring speech connectedness in different contexts (Bertola et al., 2014; Botezatu et al., 2022; Malcorra et al., 2021, 2022; Mota et al., 2012, 2016, 2017, 2018, 2020, 2022, 2023; Palaniyappan et al., 2019; Pinheiro et al., 2020). By objectively assessing the narrative complexity level of an oral text, it becomes possible to align reading materials to an individual's ZPD. This method represents words as nodes and connects consecutive words in a narrative through directed edges, creating a word-recurrence graph (Mota et al., 2014, 2023). Using this graph representation, we can quantify narrative structure through metrics such as long-range recurrences (e.g., the number of nodes within the largest strongly connected component, or LSC) and short-range recurrences (e.g., the number of repeated edges) (Mota et al., 2018). Studies show that oral narrative complexity increases as students progress through school. This development is marked by a gradual increase in long-range recurrences, which stabilize by the end of high school, and a sharp, steady decline in short-range recurrences, evidenced by a significant reduction in repeated word associations (Mota et al., 2018). These changes become evident as soon as children start to read, reflecting the rapid impact of literacy acquisition on narrative structure (Malcorra et al., 2024). This dynamic highlights how narrative complexity evolves over time, supporting its potential use for tailoring educational materials to different developmental stages.

A similar pattern of narrative evolution has been observed in the historical development of literary texts. Analyses of ancient writing - from early narratives of Sumeria and ancient Egypt to contemporary literature - revealed increased structural complexity over time. This progression is characterized by decreasing short-range and increasing long-range recurrences, with maturation and stabilization of complexity evident in classic texts from the Greco-Roman tradition (Pinheiro et al., 2020).

Given this historical trend, we focus on the progression of narrative complexity in literary texts used across different grade levels in education. By applying word-recurrence graph analysis, we hypothesize that long-range recurrences will increase in a way that is similar to historical trends. In contrast, short-range recurrences will decrease as the school year for which a book is deemed appropriate advances. This hypothesis aligns with the idea that educational texts are intentionally structured to align with learners' cognitive and linguistic development at each stage.

To assess the narrative complexity of literary texts used in schools, we analyzed a dataset of 1,627 texts through word-recurrence graph analysis. The recommended grade levels categorized the texts, and we calculated the evolution of their connectedness concerning these classifications. This approach allowed us to examine how connectedness develops across various educational stages. Our findings reveal a significant exponential increase in connectedness, particularly during the first three years of education, supporting our hypothesis. These results indicate that narrative complexity systematically increases in books designed for older children. For instance, variations in the *Cinderella* tale, tailored for first and fifth graders in elementary school, demonstrate this progression (Figure 1C). Insights into narrative complexity can inform more effective reading recommendations for children, fostering improved literacy skills and encouraging lifelong reading habits. Furthermore, maintaining lifelong reading and writing habits can contribute to cognitive reserve and offset the effects of aging on speech structure in older adults, as evidenced in a word-recurrence graph analysis (Malcorra et al., 2022).

Significant Spearman correlations were observed between recommended years and the seven analyzed graph attributes ($p < 0.001$). Positive correlations were found for nodes (N; Rho = 0.20), edges (E; Rho = 0.54), long-range recurrence attributes - largest connected component (LCC; Rho = 0.43) and largest strongly connected component (LSC; Rho = 0.51) - and graph size (ASP; Rho = 0.28). Conversely, negative correlations were observed for short-range recurrence attributes - repeated edges (RE; Rho = -0.12) and parallel edges (PE; Rho = -0.09). These findings align with expectations, indicating that more complex text - characterized by fewer repetitions and greater narrative complexity - is recommended for more advanced and fluent students. This supports the idea that narrative complexity increases systematically with the grade level for which a text is intended.

Long-range recurrence attributes consistently increased across the recommended grade levels, while short-range recurrence attributes decreased, reflecting the growing narrative complexity of literary texts. Notably, short-range recurrence attributes (RE and PE) appeared to have a smaller influence on the overall dynamics, potentially due to their spontaneous and oral nature, as suggested by their smaller effect sizes compared to long-range recurrence attributes. This developmental dynamic follows an asymptotic curve, providing insights into the optimal timing for targeting specific attributes in reading curriculum design. Our findings show short recurrence attributes mature around Year 2, nodes and ASP by Year 3, edges and LCC by Year 4, and LSC by Year 5 (Figure 1D). This progression aligns with previously documented increases in oral narrative complexity (Mota et al., 2023), further emphasizing the value of literary texts in fostering literacy skills.

Our results demonstrate the potential of using graph-theoretical attributes to recommend reading materials and tailor personalized reading curricula, particularly during critical early literacy years. Word-recurrence graph analysis offers a quantitative and automated method to assist specialists - such as publishers and librarians - in determining the suitability of reading materials for specific grade levels. The observed exponential growth in narrative complexity within literary texts reinforces their role as a powerful tool for improving reading fluency and literacy.

In conclusion, this study highlights the dynamic progression of narrative complexity in literary texts used in educational settings. These findings promise a foundation for practical reading recommendations and personalized curricula, offering educators valuable tools to enhance literacy skills and promote lifelong reading habits in young learners.

**Methods Section**

*Selection of the collection and pre-processing*

The dataset analyzed in this study comprised 1,627 written texts used in a reading curriculum generously provided by *Árvore*, an educational startup offering a digital reading application. Written in plain Portuguese, these texts included metadata about the recommended grade level for each text. Librarians manually determined grade level recommendations using a framework that aligns cognitive developmental stages with reading development milestones (Silva et al., 2010). Year 0 corresponds to kindergarten, Years 1–9 corresponds to primary education, Year 10 corresponds to high school, and Year 11 corresponds to literary content suitable for adults. To ensure the focus remained on narrative structure, only texts written in prose were included in the analysis. Pre-processing was performed using a Python script employing regular expressions to remove editorial elements. This routine eliminated catalog cards, indexes, tables of contents, bibliographic references, and other non-narrative components while preserving the line breaks in the text. Retaining line breaks was essential, as word-graph construction treats each line break as a boundary for a new graph. Moreover, line breaks were particularly relevant for texts intended for younger readers, where formatting often included breaks within paragraphs.

*Graph analysis*

The narrative dynamics of the texts were analyzed using word-recurrence graphs created with the *Strix* software. In these graphs, nodes represent individual words, and edges represent the linear sequence between words (Figure 1A). Given the substantial size of the literacy texts, the analysis employed a moving window approach to calculate average graph attributes. This involved a window of 30 words with 50% overlap (i.e., steps of 15 words) and calculating graph attributes for each resulting graph (Figure 1B). The analysis focused on seven graph attributes: (a) the number of nodes (N), a proxy of lexical diversity; (b) the number of edges (E), defined as the connections between nodes; (c) the number of repeated edges (RE), defined as the sum of all edges linking the same pair of nodes; (d) the number of parallel edges (PE), defined as the total number of edges connecting the same pair of words more than once; (e) the number of nodes in the largest connected component (LCC), defined as the largest set of nodes directly or indirectly linked by some path; (f) the number of nodes in the largest strongly connected component (LSC), defined as the largest set of nodes directly or indirectly linked by reciprocal paths, so that all the nodes in the component are mutually reachable; and (g) the average shortest path (ASP), a measure of the graph size, defined as the average distance of the shortest path between pairs of nodes in a network, calculated by the sum of all shortest paths divided by the number of paths (Mota et al., 2012, 2014, 2023). In word-recurrence graphs, relationships between nodes are defined by the sequence of words rather than their semantic or linguistic relationships. This ensures that the graph attributes reflect patterns of word recurrence independent of meaning (Mota et al., 2023).

*Statistical analyses*

The relationship between the recommended grade level and the graph attributes was assessed using Spearman's correlations. To gain deeper insights into the narrative dynamics observed in the data, the following asymptotic exponential model was employed: $f(t) = f_0 + (f_\infty - f_0)(1 - exp(-t/T))$ where $f_\infty$ represents the maximum asymptotic graph attribute value, $f_0$ denotes the initial graph attribute value, *t* is time represented as recommended years, and *T* is the characteristic time required to reach saturation, as described by Pinheiro et al. (2020). All the analyses were conducted with Python 3.9.16.

Figure 1: **The narrative complexity in textbooks increases with the school grade level. A.** Word graphs are a method for analyzing the structural complexity of texts. Text excerpts are transcribed, with each word represented as a node and the sequence between them as directed edges. **B.** Each text window contains 30 words, and graph attributes are computed for each sequence of 30 words with a 50% overlap. Analyses are based on the average attributes across windows. **C.** For example, the *Cinderella* text is used at different grade levels. A text window discussing the same theme, with the same number of words, exhibits a more complex narrative structure in grade 5 compared to kindergarten. **D.** The asymptotic model more accurately represents the development of narrative structural complexity in education.

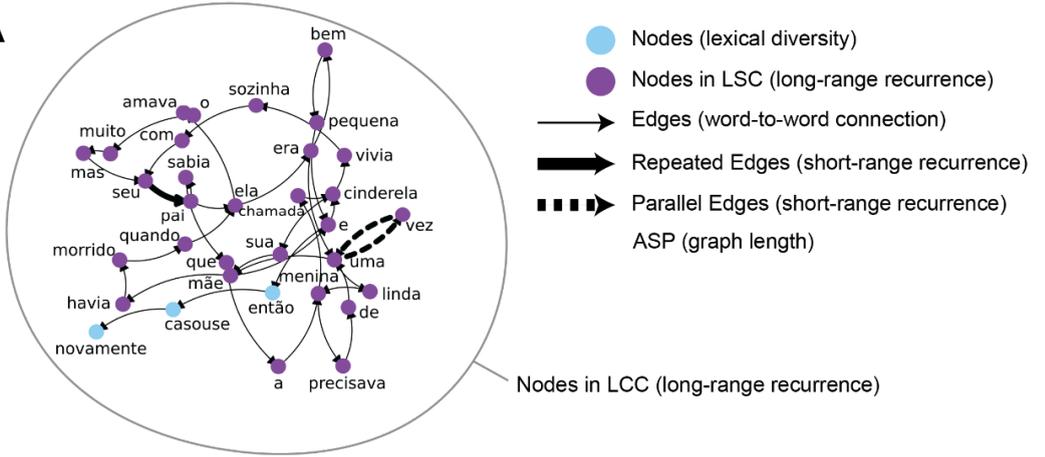
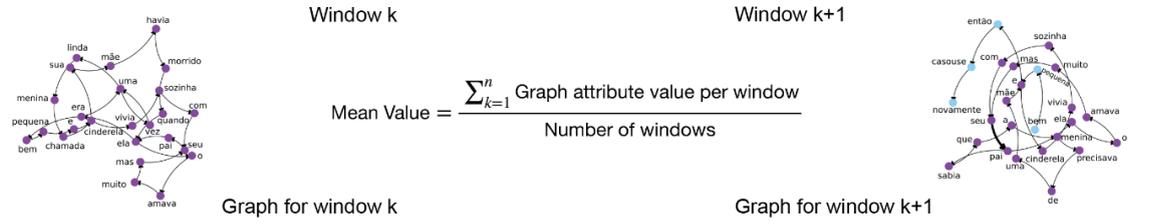
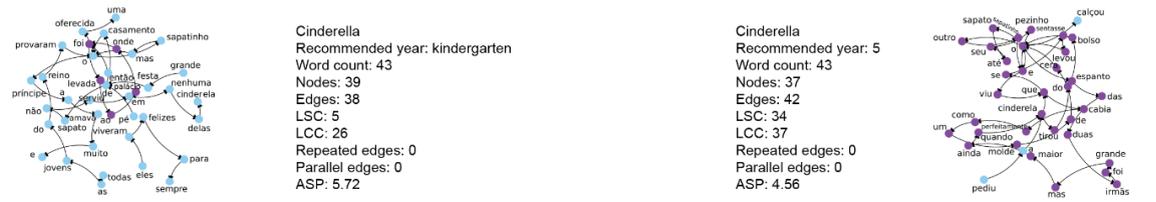
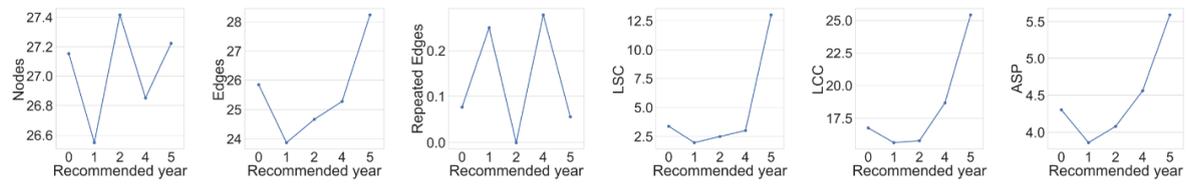
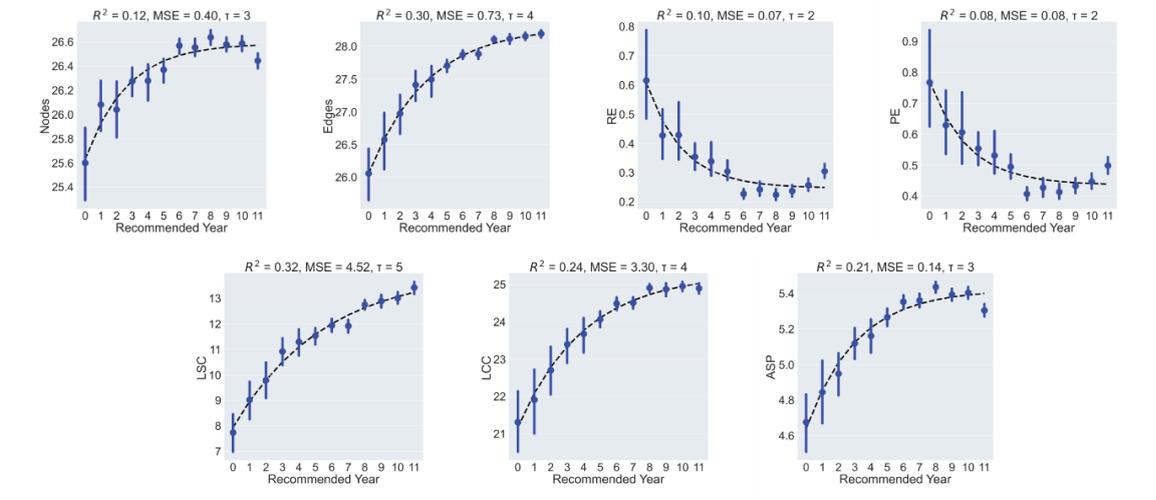

Table 1: **Statistical parameters considered in the analysis.** Spearman correlation values and p-values (alpha = 0.007 for 7 comparisons) for the relationship between graph attributes and recommended school years. The asymptotic model provided the best representative fit between graph attributes and recommended years. The table presents $R^2$, MSE (Mean Squared Error), and the parameters of the asymptotic model. $\tau$ represents the characteristic year of graph attribute saturation, $f_0$ the initial attribute value, and $f_\infty$ its maximum asymptotic value.

|  | N | E | RE | PE | LSC | LCC | ASP |
|---|---|---|---|---|---|---|---|
| **Spearman $\rho$** | 0.20 | 0.54 | -0.12 | -0.09 | 0.51 | 0.43 | 0.28 |
| **p-value** | p < 0.001 | p < 0.001 | p < 0.001 | p < 0.001 | p < 0.001 | p < 0.001 | p < 0.001 |
| **Asymptotic $R^2$** | 0.12 | 0.30 | 0.10 | 0.08 | 0.32 | 0.24 | 0.21 |
| **Asymptotic MSE** | 0.40 | 0.72 | 0.07 | 0.08 | 4.5 | 3.3 | 0.14 |
| **Asymptotic $\tau$** | 3 | 4 | 2 | 2 | 5 | 4 | 3 |
| **Asymptotic $f_0$** | 26 | 26 | 1 | 1 | 8 | 21 | 5 |
| **Asymptotic $f_\infty$** | 27 | 28 | 0 | 0 | 14 | 25 | 5 |

**Data Availability**

The datasets used and/or analyzed during the current study are available from the corresponding author upon reasonable request.

**Acknowledgments**

We want to acknowledge all professionals from Árvore who directly or indirectly contributed to the success of this project.

**Funding**

The study received private funds from Mobile Brain S.A.

**Competing Interests**

M.R., B.M. and N.B.M. work at Mobile Brain S.A., an EdTech startup. D.P. worked at Árvore, an EdTech startup, during the manuscript development.

**Author Contributions**

M.R., B.M. and N.B.M. conceived and supervised the project. D.P. curated the dataset. M.R. preprocessed and analyzed the data. M.R., B.M., and N.B.M. wrote the manuscript. All the authors read and approved the final version of the manuscript.